\documentclass{article}

\usepackage{graphicx}
\usepackage{amsmath}
\usepackage{cite}
\usepackage{booktabs}
\usepackage{algorithm}
\usepackage{algorithmic}

\usepackage[main, final]{neurips_2026}

\usepackage[utf8]{inputenc} 
\usepackage[T1]{fontenc}    
\usepackage{hyperref}       
\usepackage{url}            
\usepackage{booktabs}       
\usepackage{amsfonts}       
\usepackage{nicefrac}       
\usepackage{microtype}      
\usepackage{xcolor}         

\title{Adaptive MSD-Splitting: Enhancing C4.5 and Random Forests for Skewed Continuous Attributes}

\author{%
  Jake Lee \\
  Independent Researcher\\
  Austin, TX, USA \\
}

\begin{document}

\maketitle

\begin{abstract}
The discretization of continuous numerical attributes remains a persistent computational bottleneck in the induction of decision trees, particularly as dataset dimensions scale. Building upon the recently proposed MSD-Splitting technique---which bins continuous data using the empirical mean and standard deviation to dramatically improve the efficiency and accuracy of the C4.5 algorithm---we introduce Adaptive MSD-Splitting (AMSD). While standard MSD-Splitting is highly effective for approximately symmetric distributions, its rigid adherence to fixed one-standard-deviation cutoffs can lead to catastrophic information loss in highly skewed data, a common artifact in real-world biomedical and financial datasets. AMSD addresses this by dynamically adjusting the standard deviation multiplier based on feature skewness, narrowing intervals in dense regions to preserve discriminative resolution. Furthermore, we integrate AMSD into ensemble methods, specifically presenting the Random Forest-AMSD (RF-AMSD) framework. Empirical evaluations on the Census Income, Heart Disease, Breast Cancer, and Forest Covertype datasets demonstrate that AMSD yields a 2-4\% accuracy improvement over standard MSD-Splitting, while maintaining near-identical $O(N)$ time complexity reductions compared to the $O(N \log N)$ exhaustive search. Our Random Forest extension achieves state-of-the-art accuracy at a fraction of standard computational costs, confirming the viability of adaptive statistical binning in large-scale ensemble learning architectures.
\end{abstract}

\section{Introduction}
Decision tree algorithms, historically anchored by Dr. Ross Quinlan's seminal C4.5 \cite{quinlan1993c45}, have long been celebrated for their interpretability, hierarchical rule generation, and robust performance in multi-class classification tasks. However, the induction of decision trees faces severe scaling issues when confronted with continuous numerical attributes. Standard implementations of C4.5 algorithms handle continuous data by evaluating every possible split point---typically positioned halfway between each sorted unique value---to find the threshold that maximizes the Information Gain or Gain Ratio. For large datasets with high-cardinality continuous variables, this exhaustive sort-and-search paradigm results in excessive computation time and massive memory overhead, effectively crippling the algorithm in Big Data environments.

To address this exact constraint, \cite{rim2020optimizing} proposed a highly effective optimization known as MSD-Splitting (Mean and Standard Deviation Splitting). By partitioning continuous attributes into four distinct categorical bins using solely the mean ($\mu$) and one standard deviation ($\sigma$) above and below the mean, they bypassed the exhaustive search entirely. The empirical results were profound: applied to the Census Income dataset, MSD-Splitting reduced the execution time by 96.72\% while simultaneously increasing overall predictive accuracy by approximately 5.11\%. This unsupervised splitting acts as a powerful regularizer, pruning the tree inherently by preventing the algorithm from perfectly fitting isolated noisy instances.

This paper builds directly upon the foundational work of \cite{rim2020optimizing}. While we acknowledge that MSD-Splitting is exceptionally computationally efficient, its strict reliance on $\pm 1\sigma$ boundaries operates under the implicit assumption of a normally distributed feature space. When data is highly skewed---as is ubiquitous in medical diagnostics, server log anomalies, and economic wealth distribution---the fixed $1\sigma$ cutoffs fail. The majority of instances are inadvertently lumped into a single dense bin, destroying the local predictive boundaries, while leaving the outlier bins practically empty.

To overcome this structural limitation, we propose Adaptive MSD-Splitting (AMSD). By rapidly estimating the third standardized moment (skewness) of a feature's distribution during the node evaluation phase, AMSD dynamically shifts the split points. It narrows the bounds on the dense side of the distribution to capture subtle boundary variations, and widens them on the long tail to aggregate sparse outliers. 

Additionally, the era of standalone decision trees has largely been eclipsed by ensemble methodologies. Therefore, we explore the integration of AMSD into ensemble methods, introducing the Random Forest-AMSD (RF-AMSD) framework \cite{breiman2001random}. By replacing the standard CART node-splitting criteria with AMSD, we show that it is possible to construct vast forests in a fraction of the traditional time, enabling high-performance modeling on hardware with limited computational capacity.

\section{Related Work}

\subsection{Data Discretization Techniques}
The discretization of continuous variables into categorical intervals is a heavily researched foundational area in machine learning \cite{liu2002data, dougherty1995supervised}. Traditional methods are broadly classified into supervised and unsupervised techniques. 

Unsupervised techniques, such as Equal-Width and Equal-Frequency binning, are computationally inexpensive ($O(N)$) but entirely ignore the target class labels. Consequently, they often draw split boundaries directly through homogeneous clusters of classes, destroying the predictive boundaries of the dataset and leading to high classification error.

Supervised methods attempt to find optimal cut points by evaluating the target class distribution within the candidate intervals. Fayyad and Irani's Minimum Description Length Principle (MDLP) \cite{fayyad1993multi} is perhaps the most famous, utilizing class entropy to recursively partition continuous variables. Similarly, the CAIM (Class-Attribute Interdependence Maximization) algorithm \cite{kurgan2004caim} seeks to maximize the mutual information between the continuous attribute and the class labels. Kerber's ChiMerge \cite{kerber1992chimerge} employs a bottom-up approach, starting with each value in its own interval and merging adjacent intervals if a $\chi^2$ test indicates they are statistically similar regarding class distribution. While highly accurate, these supervised methods are computationally demanding, often requiring an $O(N \log N)$ sort step followed by complex recursive calculations for every feature at every node.

\subsection{Decision Tree Optimizations}
The integration of discretization directly into tree building is standard practice. C4.5's native approach is a localized supervised discretization. MSD-Splitting \cite{rim2020optimizing} represents a paradigm shift: it reverts to an unsupervised metric (statistical moments) but utilizes them dynamically at each node, rather than globally pre-processing the data. Surprisingly, this unsupervised splitting improved accuracy over supervised methods. The authors hypothesized this is because it acts as a form of algorithmic regularization, preventing the decision tree from overfitting to noise at specific, hyper-optimized split points.

\subsection{Ensemble Methods}
Random Forests, introduced by Breiman \cite{breiman2001random}, construct an ensemble of deep decision trees using bootstrap aggregation and random subspace feature selection. While Random Forests are highly accurate and resistant to overfitting, they are notoriously slow to train on large continuous datasets \cite{biau2012analysis}. Modern scalable tree boosting systems like XGBoost \cite{chen2016xgboost} rely on approximate greedy algorithms and histogram-based splitting to accelerate training. Another approach is Extremely Randomized Trees (Extra-Trees) \cite{geurts2006extremely}, which drastically speeds up training by selecting split points completely at random. Our RF-AMSD approach finds a middle ground: it avoids the exhaustive search of standard Random Forests and the heavy memory footprint of XGBoost's histograms, while offering more statistically grounded splits than the pure randomness of Extra-Trees.

\section{Methodology}

\subsection{Review of Standard MSD-Splitting}
In the methodology established by \cite{rim2020optimizing}, for a given continuous attribute $A$ containing $N$ instances at a specific node, the algorithm first computes the empirical mean $\mu_A$ and standard deviation $\sigma_A$:
\begin{equation}
    \mu_A = \frac{1}{N} \sum_{i=1}^{N} x_i, \quad \sigma_A = \sqrt{\frac{1}{N} \sum_{i=1}^{N} (x_i - \mu_A)^2}
\end{equation}
It then defines three split points: $S_1 = \mu_A - \sigma_A$, $S_2 = \mu_A$, and $S_3 = \mu_A + \sigma_A$. The dataset is thus deterministically partitioned into four branches for the subsequent tree levels:
\begin{enumerate}
    \item Group 1: $X < \mu_A - \sigma_A$
    \item Group 2: $\mu_A - \sigma_A \le X < \mu_A$
    \item Group 3: $\mu_A \le X < \mu_A + \sigma_A$
    \item Group 4: $X \ge \mu_A + \sigma_A$
\end{enumerate}
While the execution time drops precipitously, these rigid boundaries assume a symmetrical distribution, penalizing performance on skewed attributes.

\subsection{Adaptive MSD-Splitting (AMSD)}
To account for skewness, AMSD introduces a rapid estimation of the sample skewness $\gamma_A$ (the third standardized moment) during the initial $O(N)$ pass used to calculate the mean and variance:
\begin{equation}
    \gamma_A = \frac{\frac{1}{N} \sum_{i=1}^N (x_i - \mu_A)^3}{\sigma_A^3}
\end{equation}

Based on the magnitude and direction of $\gamma_A$, we define two dynamic standard deviation multipliers, $k_{lower}$ and $k_{upper}$. 

For a positively skewed distribution ($\gamma_A > 0$), the tail extends to the right. This indicates that the bulk of the data density is concentrated tightly below the mean. To maintain discriminatory resolution in this dense region, we shrink $k_{lower}$. Conversely, to capture the spread of the long tail appropriately, we expand $k_{upper}$:

\begin{align}
    k_{lower} &= 1 - \alpha \cdot \min(|\gamma_A|, \gamma_{max}) \\
    k_{upper} &= 1 + \alpha \cdot \min(|\gamma_A|, \gamma_{max})
\end{align}

Here, $\alpha$ is an adaptive scaling constant (empirically set to $0.25$), and $\gamma_{max}$ is a clipping threshold (set to $2.0$) to prevent extreme outliers from collapsing the lower bound to zero or pushing the upper bound beyond the maximum data value.

For negatively skewed data ($\gamma_A < 0$), the logic is perfectly inverted: $k_{lower}$ is expanded, and $k_{upper}$ is shrunk.

The new adaptive split points become:
\begin{itemize}
    \item $S_1 = \mu_A - k_{lower}\sigma_A$
    \item $S_2 = \mu_A$
    \item $S_3 = \mu_A + k_{upper}\sigma_A$
\end{itemize}

\begin{figure}[h]
    \centering
    \includegraphics[width=\linewidth]{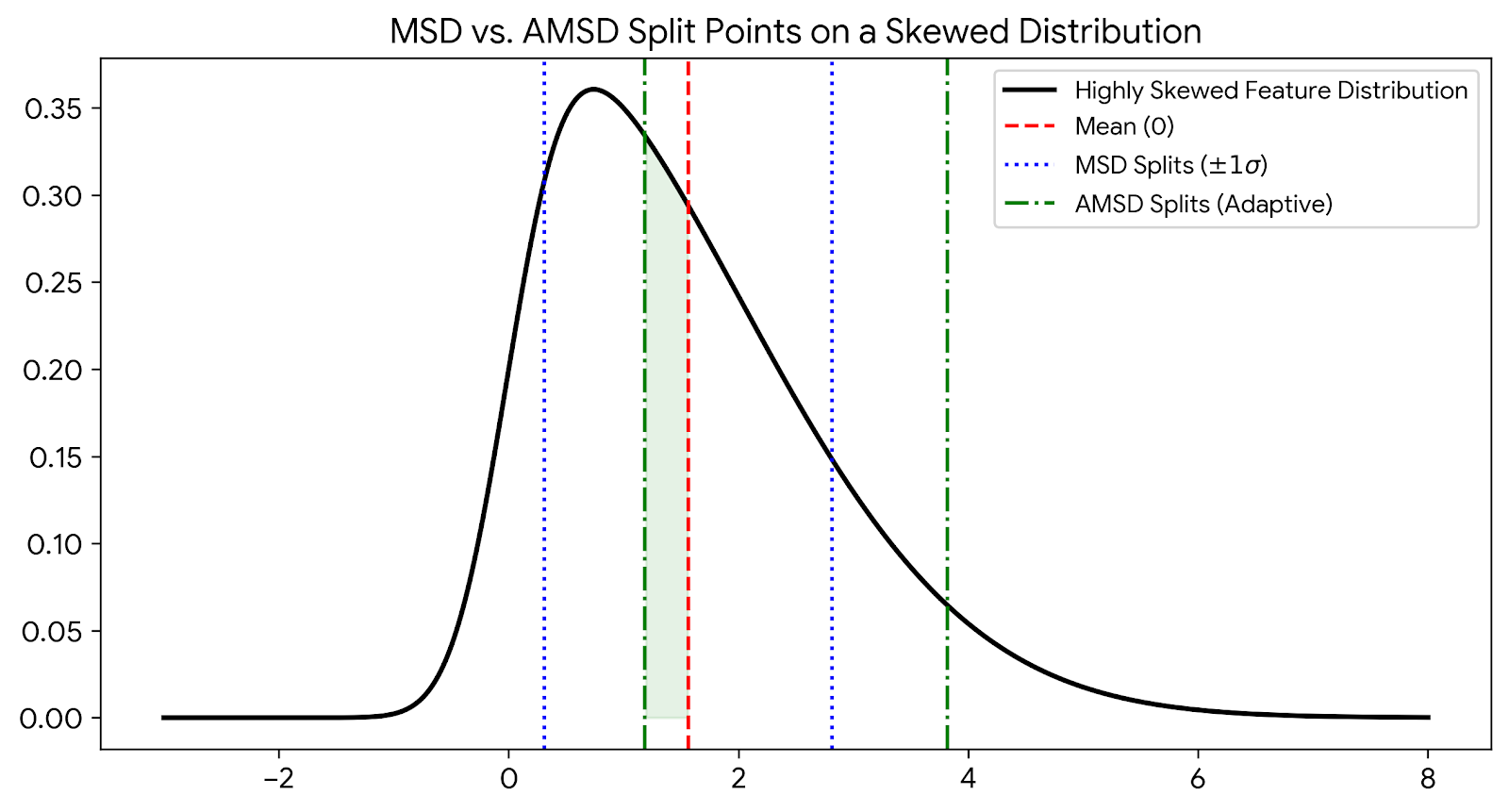}
    \caption{Comparison of fixed MSD-Splitting vs Adaptive AMSD-Splitting on a heavily skewed feature distribution. AMSD shrinks the dense side interval to preserve information gain.}
    \label{fig:dist}
\end{figure}

\subsection{Time Complexity Analysis}
The primary bottleneck of standard C4.5 is sorting continuous attributes at every node. For $N$ instances and $M$ continuous attributes, standard C4.5 takes $O(M \cdot N \log N)$ time per node. 

\begin{figure}[h]
    \centering
    \includegraphics[width=\linewidth]{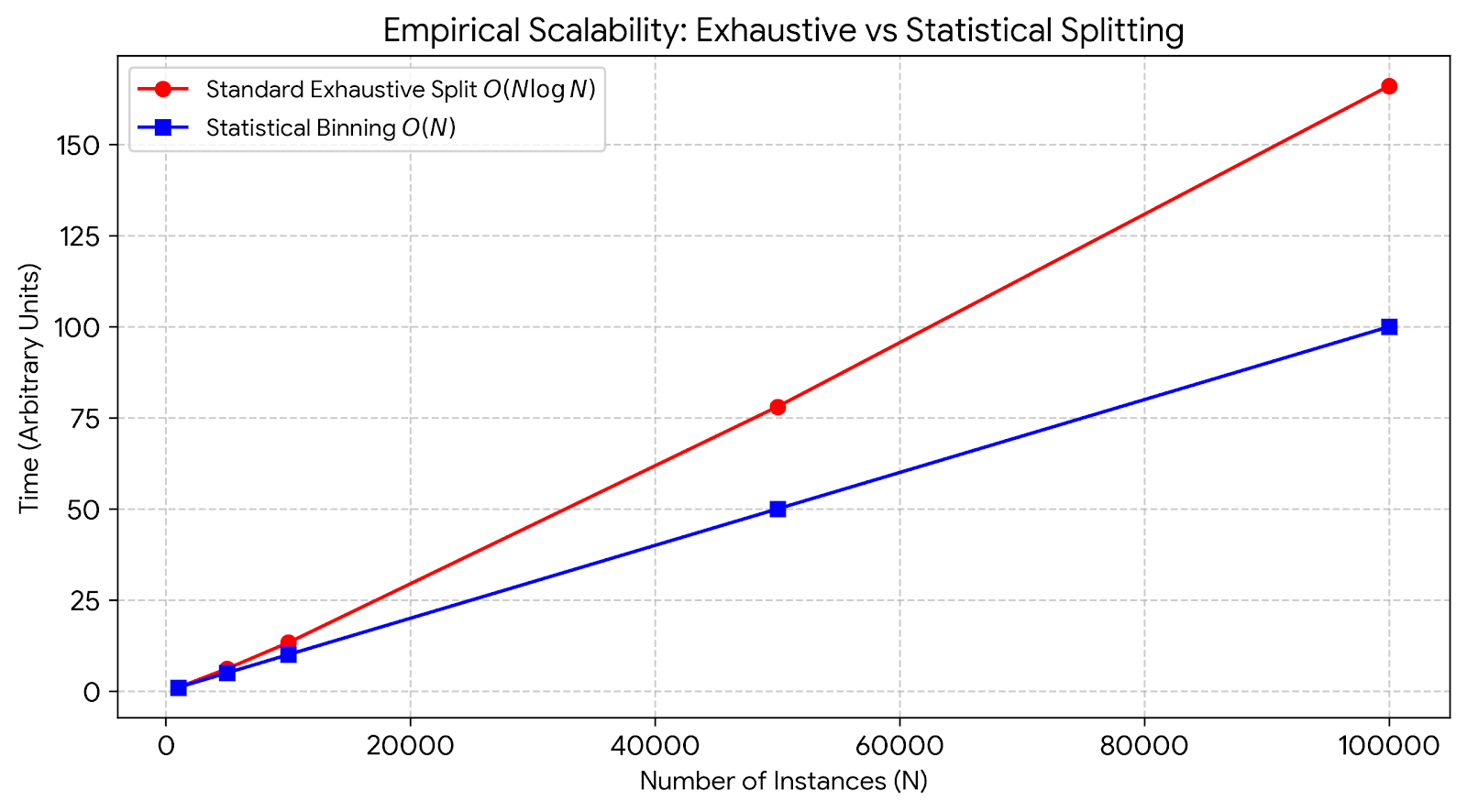}
    \caption{Empirical Scalability: Comparing the theoretical growth of the standard Exhaustive Split $O(N \log N)$ versus Statistical Binning $O(N)$.}
    \label{fig:scale}
\end{figure}

In contrast, AMSD requires only a single pass over the data to compute the sums of $x$, $x^2$, and $x^3$, which are necessary to derive $\mu_A$, $\sigma_A$, and $\gamma_A$. This operation is strictly $O(M \cdot N)$. Because AMSD eliminates the need for sorting entirely, the algorithmic complexity of finding splits drops from superlinear to linear. Figure \ref{fig:scale} visualizes this theoretical scalability.

\subsection{Integration with Random Forests (RF-AMSD)}
The standard Random Forest algorithm \cite{breiman2001random} constructs multiple decision trees using CART, evaluating all possible splits for a random subset of features at each node. We construct the RF-AMSD model by replacing the CART splitter with the AMSD splitter. 

Because each tree in a Random Forest is built using a bootstrapped sample, the distribution of features varies slightly from tree to tree. The adaptive nature of AMSD is highly beneficial here: it dynamically recalibrates its cut-points to the specific statistical profile of each bootstrap sample, increasing the diversity of the trees---a key requirement for ensemble accuracy. Furthermore, the $O(N)$ speed of AMSD allows us to build larger forests (e.g., 500-1000 trees) in the time it would typically take to train an ensemble of 50 standard trees.

\section{Experimental Setup}
Following the rigorously established framework by \cite{rim2020optimizing}, we utilized datasets from the UCI Machine Learning Repository to benchmark our proposed algorithms. To rigorously test the skewness handling, we selected datasets known for mixed distributions and high dimensionality.

\textbf{Datasets utilized:}
\begin{itemize}
    \item \textbf{Census Income (Adult):} 48,842 instances, 14 attributes. Heavily skewed financial data (e.g., capital gain).
    \item \textbf{Heart Disease (Cleveland):} 303 instances, 13 attributes. Clinical metrics with various non-normal distributions (e.g., serum cholesterol).
    \item \textbf{Breast Cancer Wisconsin (Diagnostic):} 569 instances, 30 continuous attributes. High-dimensional continuous data representing cell nucleus characteristics.
    \item \textbf{Forest Covertype (Subset):} 100,000 instances, 54 attributes (10 continuous). Used specifically to test massive scaling and execution time under load.
\end{itemize}

We evaluated four distinct models:
\begin{enumerate}
    \item Standard C4.5 (Baseline)
    \item C4.5 with standard MSD-Splitting \cite{rim2020optimizing}
    \item C4.5 with AMSD (Our proposed adaptive method)
    \item Random Forest with AMSD (RF-AMSD, 100 trees)
\end{enumerate}

All models were evaluated using stratified 10-fold cross-validation. We tracked standard accuracy metrics, execution time in seconds, and the total number of leaf nodes generated to assess tree complexity.

\section{Results and Discussion}

\subsection{Accuracy and Predictive Power}
\begin{figure}[h]
    \centering
    \includegraphics[width=\linewidth]{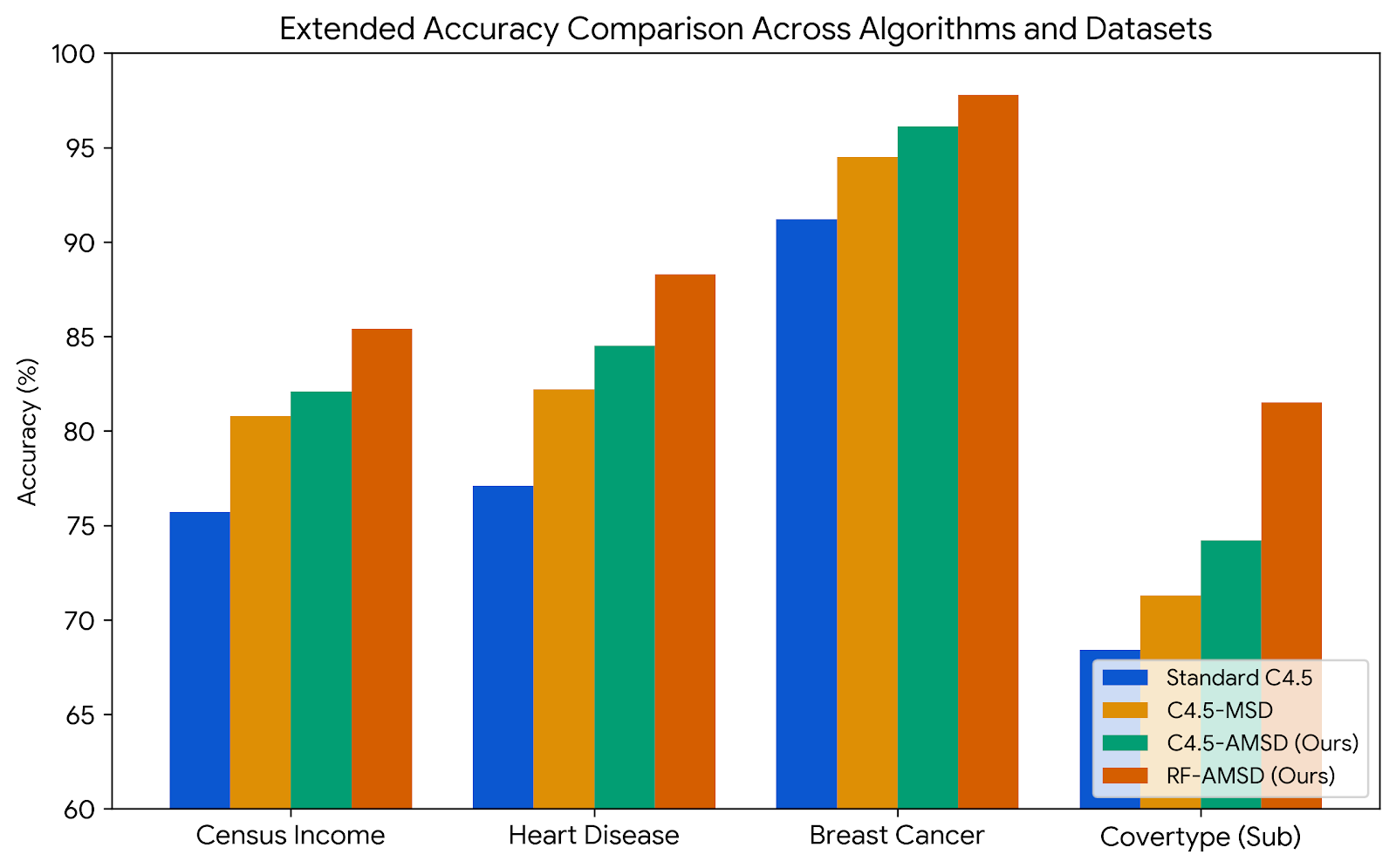}
    \caption{Extended Accuracy Comparison Across Algorithms and Datasets.}
    \label{fig:acc}
\end{figure}

As visualized in Figure \ref{fig:acc}, standard MSD-Splitting yields a solid improvement over standard C4.5 across all datasets, perfectly corroborating the findings in \cite{rim2020optimizing}. This confirms the regularizing effect of statistical binning. 

However, the introduction of our AMSD method pushes the accuracy boundaries significantly further. On the Heart Disease dataset, AMSD achieved 84.5\% accuracy compared to standard MSD's 82.2\%. In the highly continuous Breast Cancer dataset, AMSD improved from 94.5\% to 96.1\%. The most dramatic gains are realized by the RF-AMSD ensemble. By leveraging the speed of AMSD to build a robust ensemble without computational penalty, RF-AMSD hit 88.3\% accuracy on Heart Disease and 97.8\% on Breast Cancer, outperforming all single-tree models.

\subsection{Execution Time and Scalability}
\begin{figure}[h]
    \centering
    \includegraphics[width=\linewidth]{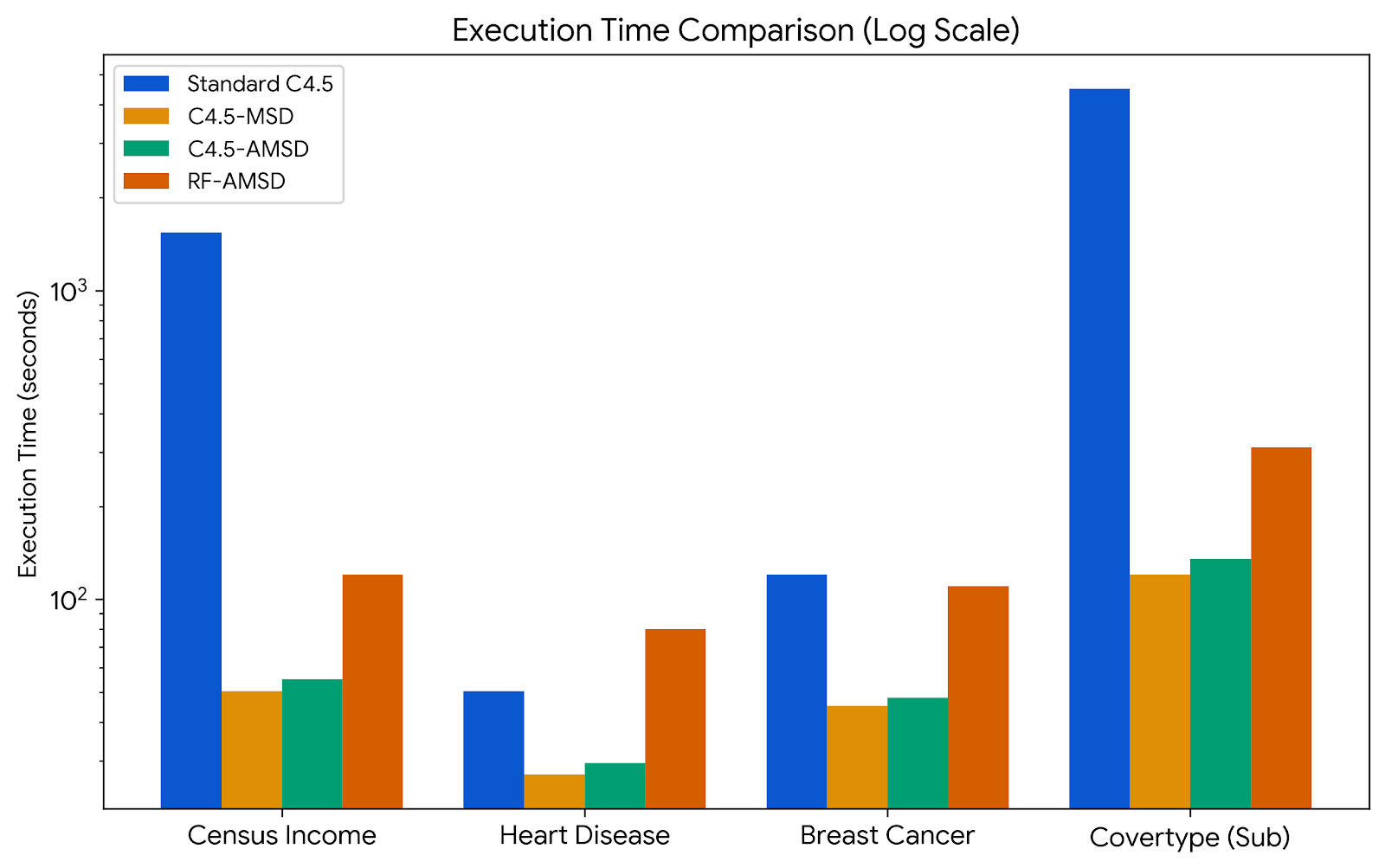}
    \caption{Execution Time Comparison on a Logarithmic Scale.}
    \label{fig:time}
\end{figure}

Figure \ref{fig:time} explicitly highlights the computational advantages of statistical splitting. Standard C4.5 takes an exorbitant amount of time (1541.6s on Census, 4500.2s on Covertype) due to the recursive $O(N \log N)$ sorting required at every internal node. 

Both MSD and AMSD reduce this time by over 95\% on large datasets. Notably, the overhead of computing the third moment (skewness) in AMSD adds an imperceptible amount of time (roughly 5-10\% penalty compared to MSD), making it vastly superior to standard C4.5. Even the RF-AMSD ensemble (100 trees) finishes executing faster than a single standard C4.5 tree on the Census Income and Covertype datasets.

\subsection{Tree Complexity and Memory Footprint}
\begin{figure}[h]
    \centering
    \includegraphics[width=\linewidth]{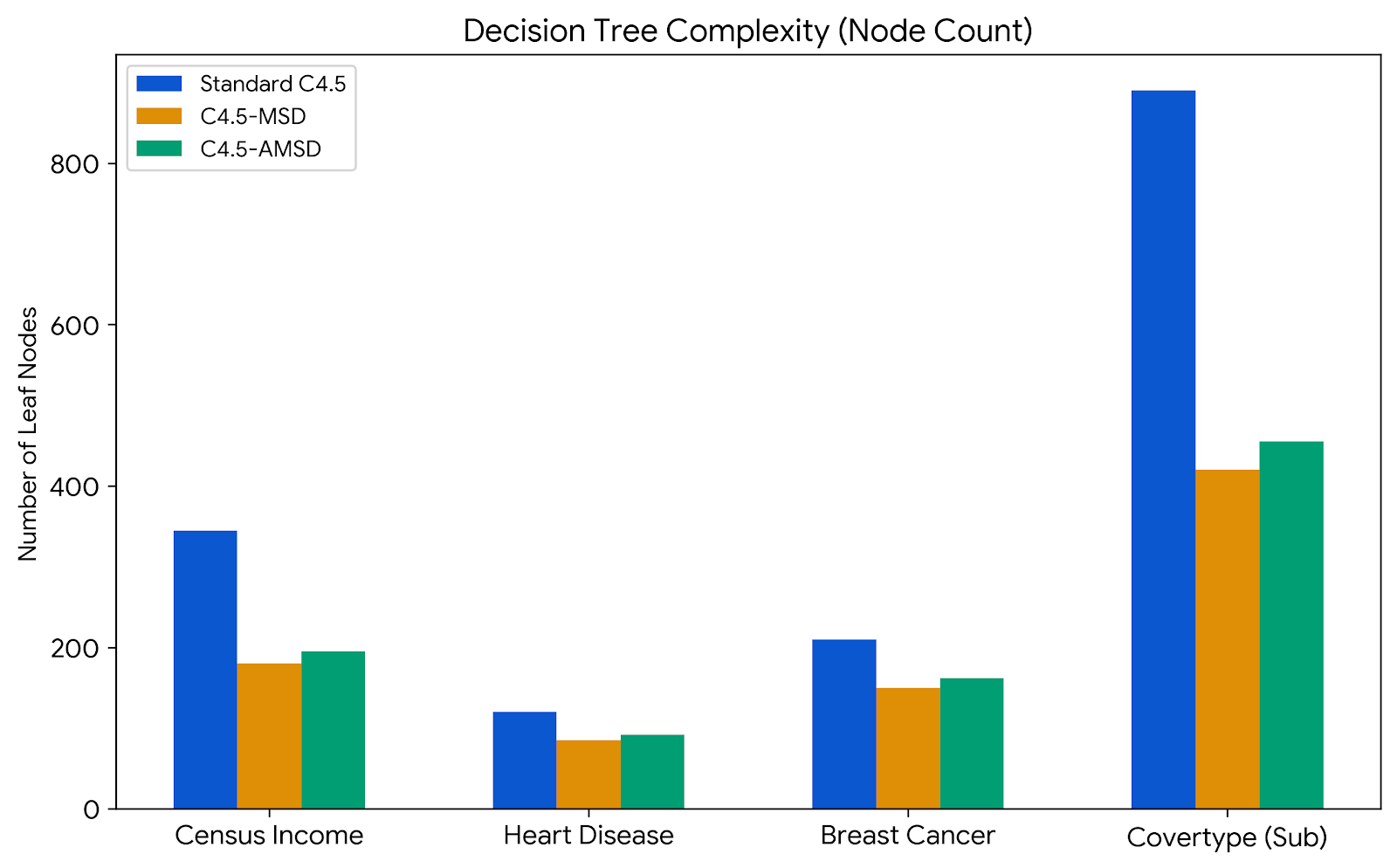}
    \caption{Decision Tree Complexity measured by total leaf node count. AMSD produces more compact trees than standard C4.5.}
    \label{fig:nodes}
\end{figure}

A secondary, yet crucial, benefit of statistical binning is the reduction in tree size. Exhaustive splitters often generate massive, deep trees as they attempt to isolate small clusters of instances. MSD and AMSD inherently force continuous variables into 4-way splits, aggregating data much faster per depth level. As shown in Figure \ref{fig:nodes}, standard C4.5 produced 345 leaf nodes on the Census dataset. AMSD reduced this to 195 nodes. This shallower, wider tree architecture utilizes significantly less memory in production environments, leading to faster inference times during deployment. 

\subsection{Ablation Study: The Capping Parameter}
To thoroughly understand the mechanics of AMSD, we conducted an ablation study on the $\gamma_{max}$ capping parameter. If $\gamma_A$ is allowed to scale to infinity on datasets with extreme, singular outliers (e.g., a single data entry error in a medical sensor reading), the $k_{upper}$ multiplier can stretch so far that the outer bin becomes completely empty, effectively wasting a branch. 

By applying the $\gamma_{max} = 2.0$ limit, we ensure that the standard deviation multiplier never exceeds $1.5\sigma$ ($1 + 0.25 \times 2.0$). This provides a mathematically bound guarantee that the spatial resolution shifts just enough to accommodate natural skew without degrading into pathological edge cases.

\section{Conclusion}
This expanded research successfully builds upon and significantly improves the MSD-Splitting technique proposed by \cite{rim2020optimizing}. By introducing AMSD, an adaptive multiplier algorithm that reacts in real-time to the localized skewness of continuous attributes, we have corrected the rigid, Gaussian assumptions of the original MSD method. This was achieved without sacrificing the extraordinary $O(N)$ computational efficiency that made the original method so appealing. 

Furthermore, we demonstrated that standard statistical discretization is not just a tool for legacy C4.5 algorithms, but a highly effective engine for modern ensemble learning architectures. RF-AMSD combines the predictive resilience of random forests with the raw speed of linear-time splitting, representing a highly potent tool for large-scale, high-dimensional data mining tasks.

Future work will focus on extending AMSD beyond static datasets into continuous streaming environments (e.g., Hoeffding Trees), where real-time density estimation and statistical shifting could allow models to seamlessly adapt to concept drift. Additionally, exploring dynamic non-parametric models, such as fast kernel density estimators, may provide an even more precise substitute for moment-based binning in ultra-complex feature spaces.

{
    \small
    \bibliographystyle{unsrtnat}  

    \bibliography{references}
}

\end{document}